\documentclass[accepted]{uai2026} 
                        
\usepackage{amsmath,amsfonts,bm}

\def\eqref#1{equation~\ref{#1}}

\def\1{\bm{1}}

\DeclareMathAlphabet{\mathsfit}{\encodingdefault}{\sfdefault}{m}{sl}
\SetMathAlphabet{\mathsfit}{bold}{\encodingdefault}{\sfdefault}{bx}{n}

\usepackage[british]{babel}

\usepackage{natbib} 
\usepackage{mathtools} 
\usepackage{booktabs} 
\usepackage{tikz} 
\usepackage[table]{xcolor}
\usepackage{graphicx} 
\usepackage{subcaption}
\usepackage{array}
\usepackage{booktabs}
\usepackage{multirow}
\usepackage{verbatim}

\usepackage{xcolor, soul}
\sethlcolor{pink}

\usepackage{pifont}
\newcommand{\noPosterior}{\textcolor{red!70}{\large$\times$}}
\newcommand{\yesPosterior}{\textcolor{green!60!black}{\large$\checkmark$}}

\title{Semantic Self-Distillation for Language Model Uncertainty}

\author[1]{Edward~Phillips}
\author[1]{Sean~Wu}
\author[1]{Fredrik~K.~Gustafsson}
\author[1]{Boyan~Gao}
\author[1, 2]{David~A.~Clifton}
\affil[1]{%
    Department of Engineering Science\\
    University of Oxford
}
\affil[2]{%
    Oxford Suzhou Centre for Advanced Research\\
     University of Oxford\\
     Suzhou
}

\begin{document}
\maketitle

\begin{abstract}
  Large language models present challenges for principled uncertainty quantification, in part due to their complexity and the diversity of their outputs. Semantic dispersion, or the variance in the meaning of sampled answers, has been proposed as a useful proxy for model uncertainty, but the associated computational cost prohibits its use in latency-critical applications. We show that sampled semantic distributions can be distilled into lightweight student models which estimate a prompt-conditioned density before the language model generates an answer token. The student model predicts a semantic distribution over possible answers; the entropy of this distribution provides a prompt-level uncertainty signal, and the probability density allows answer-level reliability evaluation. Across experiments on TriviaQA and MMLU, we find our student models perform competitively relative to the teacher's sampled semantic dispersion on a hallucination prediction task, whilst offering additional uncertainty primitives for out-of-domain detection and multiple-choice answer selection. We term this technique Semantic Self-Distillation (SSD), which can serve as a general framework for distilling predictive uncertainty in complex output spaces beyond language. 
\end{abstract}

\section{Introduction}\label{sec:intro}

\begin{figure*}[h]
    \centering
    \includegraphics[trim={0cm 10cm 7cm 0cm}, clip,width=\linewidth]{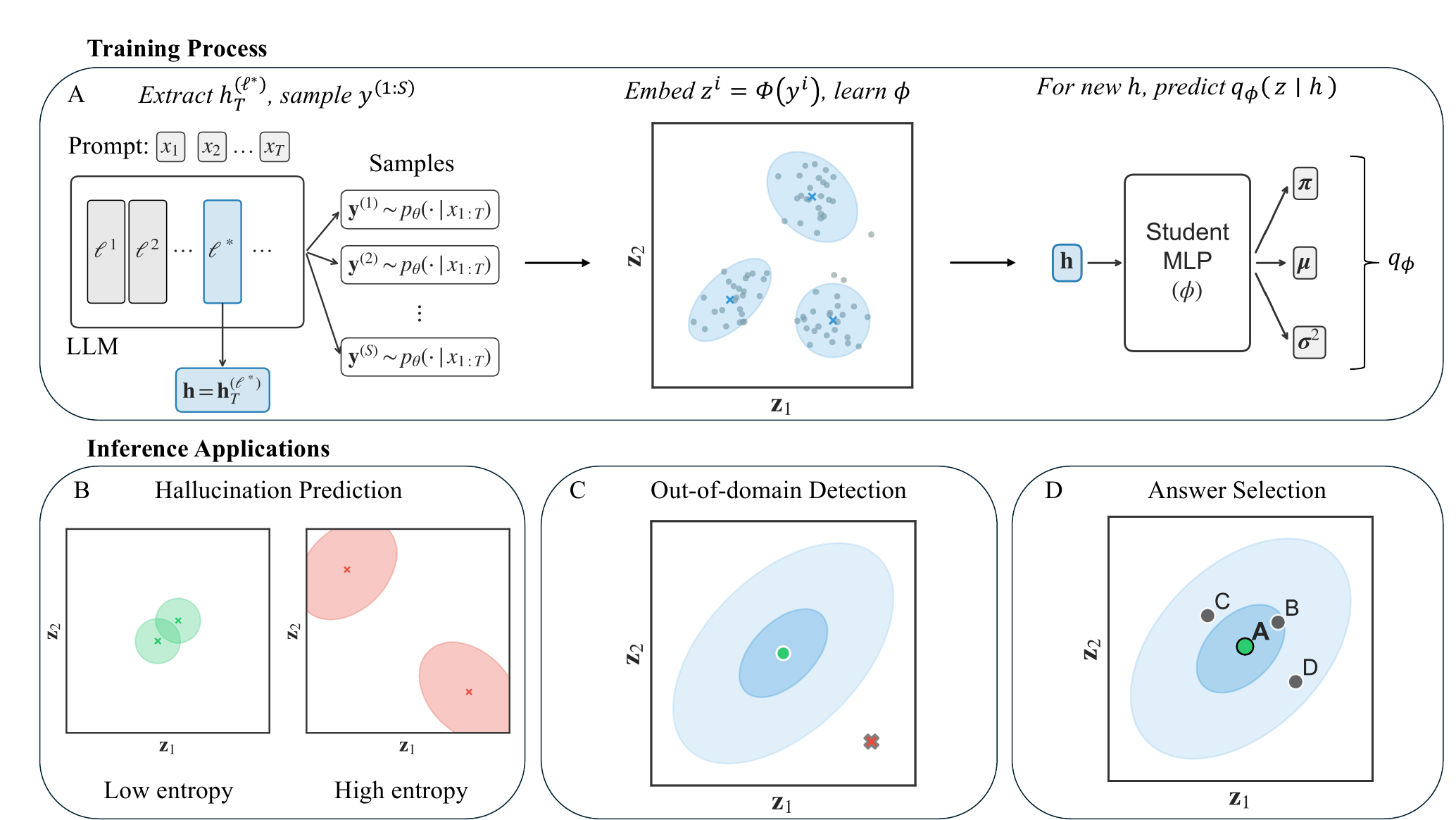}
    \caption{Overview of Semantic Self-Distillation (SSD). 
    \textbf{(A) Training:} 
    For each prompt, a hidden representation $\mathbf{h}$ conditions a student that distils sampled answer embeddings into a semantic mixture distribution.
    \textbf{(B-D) Inference:} The predicted density supports pre-generation hallucination prediction via entropy, post-generation context verification via likelihood, and multiple-choice answer selection. Plots illustrate the semantic embedding space, where points denote sampled answers.}
    \label{fig:ssd-desc}
\end{figure*}

Uncertainty quantification (UQ) can be used to assess the reliability of large language model (LLM) outputs \citep{xiong2024efficient, liullmuncertainty2025}. As LLMs are rapidly deployed in high-stakes domains such as healthcare and law, it becomes increasingly important to have meaningful measures for the trustworthiness of model-generated content. Well-designed UQ methods can provide signals for detecting errors like \textit{hallucinations} - outputs that are misaligned with fact or user intention - and enable mitigation strategies such as abstention, information retrieval, or answer selection \citep{wen-etal-2025-know}.

LLMs are now frequently embedded in \textit{agentic frameworks}, which chain multiple model invocations to solve complex, multi-step tasks \citep{yao2023react, xi2025rise}. In these settings a single unreliable output can introduce an error that compounds destructively over long-horizon tasks, potentially resulting in complete failure of the system to achieve its goal \citep{kwa2025measuring}. Given the computational burden of such agentic frameworks, practical uncertainty estimators must be low latency compared to the base language models.

\textit{Semantic dispersion} has emerged as a useful proxy for uncertainty: when a language model generates semantically diverse answers to the same prompt, the response is unlikely to be factual and correct. Sampling-based measures of semantic dispersion are effective for hallucination detection \citep{farquhar_detecting_2024}, but require generating multiple stochastic completions, rendering them prohibitively expensive in latency-sensitive applications.

Probing techniques can address this latency issue by training lightweight models to predict semantic dispersion or hallucination probability directly from internal model representations \citep{kadavath2022languagemodelsmostlyknow, obeso2025realtimedetectionhallucinatedentities, han-etal-2025-simple, kossen2024semanticentropyprobesrobust}. However, these methods typically collapse the complex notion of language model uncertainty into a single scalar metric, ignoring the topology of the underlying output distribution. This information reduction limits their application to simple risk classification, preventing the extraction of richer reliability signals.

In this work, we propose \textbf{Semantic Self-Distillation} (SSD), a general UQ framework that bridges lightweight probing techniques and expensive sampling-based methods. Instead of regressing a scalar uncertainty statistic, SSD trains a small student model to predict a prompt-conditioned distribution over semantic answer embeddings. We model the student as a mixture density network (MDN) \citep{bishop1994mixture}, and approximate the teacher distribution using stochastic samples from the base language model. The predicted density serves as a \textit{general-purpose uncertainty object}: its analytic entropy provides a pre-generation risk forecast, whilst estimated likelihoods enable post-generation answer verification. 

Figure~\ref{fig:ssd-desc} provides an overview of the SSD training procedure and the inference-time applications investigated in this work. We evaluate SSD on a hallucination prediction task with TriviaQA and MMLU across multiple LLM families, demonstrating competitive, and in some cases even superior, performance relative to the teacher dispersion baseline. 

While specialized scalar probes can be strong pre-generation classifiers, SSD uniquely retains distributional structure in a single forward pass. This enables additional post-generation capabilities unavailable to standard single-pass methods, which we demonstrate by evaluating SSD on likelihood-based out-of-domain answer detection and multiple-choice answer selection.

We summarize our contributions as the following:
\begin{itemize}
    \item We introduce \textbf{Semantic Self-Distillation} (SSD), a method for distilling a language model's sampled semantic answer distribution into a lightweight, prompt-conditioned density estimator.

    \item We show that SSD provides competitive pre-generation hallucination prediction performance across multiple model families, while matching the inference latency of single-pass probing methods.
    
    \item We demonstrate that modelling the full semantic distribution unlocks new reliability primitives beyond scalar uncertainty scoring, including post-generation answer verification and distribution-based answer selection.
\end{itemize}

\section{Methods}
\label{sec:methods}

In this section, we describe how SSD distils a language model's sampled semantic output distribution into a prompt-conditioned density estimator, rather than a scalar uncertainty proxy. We first define the representations used for prompts and answers, specify the MDN student model, and detail the distillation objective used for training. We then show how the resulting predictive density can be used to extract both pre-generation, prompt-level dispersion estimates, and post-generation, answer-level likelihood scores, all in a single forward pass at inference time.

\subsection{Problem Setup}
Let $x_{1:T}$ be an input prompt and $y_{1:L}$ a completion generated by a decoding model. Let $\mathbf{h}=f_\theta(x_{1:T})\in\mathbb{R}^{d_h}$ denote a vector representation of the prompt extracted from the model's internal states. We define a semantic representation $\mathbf{z} \in\mathbb{R}^{d_z}$ of the completed sequence via a mapping $\Phi$, which in practice is an off-the-shelf sequence embedding model:
\[
\mathbf{z}=\Phi(y_{1:L}).
\]
 We train a student model $q_\phi(\mathbf{z}\mid \mathbf{h})$ to predict the distribution over semantic embeddings $\mathbf{z}$ given only $\mathbf{h}$.

\subsubsection{Prompt and Answer Representations}
Given a user prompt $x_{1:T}$ of length $T$, we extract hidden states directly from the base language model. Let $\mathbf{h}_t^{(\ell)} \in \mathbb{R}^{d}$ denote the hidden state at layer $\ell$ for token position $t$. We utilize the representation of the \emph{final token} of the prompt extracted from a single specific layer $\ell^*$:
\[
\mathbf{h} \;=\; \mathbf{h}_T^{(\ell^*)}.
\]
The layer index $\ell^*$ is treated as a hyperparameter. In practice, we estimate the optimal $\ell^*$ by identifying the layer most predictive of the model's semantic uncertainty, as explained in Appendix~\ref{sec:layer-selection}.

For the answer, we apply a specialist embedding model $\Phi$ to the generated sequence $y_{1:L}$ to obtain the semantic representation, $\mathbf{z}=\Phi(y_{1:L})$, with full model details provided in Section \ref{sec:experiments_models_datasets}. 
To improve training efficiency and focus on the principal axes of semantic variation, we also apply Principal Component Analysis (PCA) to reduce the dimensionality of the target embeddings $\mathbf{z}$.

\subsubsection{Student Model}
We define the student model $q_\phi(\mathbf{z}\mid \mathbf{h})$ using an MDN \citep{bishop1994mixture}. 
Conditioned on the prompt representation $\mathbf{h}\in\mathbb{R}^{d_h}$, the student predicts a $K$-component Gaussian mixture over the semantic embedding $\mathbf{z}\in\mathbb{R}^{d_z}$:
$$
q_\phi(\mathbf{z}\mid \mathbf{h})
=\sum_{k=1}^{K} \pi_k(\mathbf{h})\,
\mathcal{N}\!\big(\mathbf{z};\,\boldsymbol{\mu}_k(\mathbf{h}),\,\operatorname{diag}(\boldsymbol{\sigma}_k^2(\mathbf{h}))\big),
$$
$$
\pi_k(\mathbf{h})=\mathrm{softmax}_k\big(\boldsymbol{\alpha}(\mathbf{h})\big),
$$
where an MLP with parameters $\phi$ maps $\mathbf{h}$ to mixture logits $\boldsymbol{\alpha}(\mathbf{h})\in\mathbb{R}^{K}$, component means $\boldsymbol{\mu}_k(\mathbf{h})\in\mathbb{R}^{d_z}$, and scales $\boldsymbol{\sigma}_k(\mathbf{h})\in\mathbb{R}^{d_z}$. For data efficiency, we use diagonal covariance matrices.

For each training prompt $x$, we extract the representation $\mathbf{h}$ (using the selected layer $\ell^*$) and generate $S$ stochastic completions $y^{(1)}, \dots, y^{(S)}$ using the base language model. We compute semantic embeddings $\mathbf{z}^{(s)}=\Phi(y^{(s)})$ for $s{=}1,\dots,S$, and fit the student parameters $\phi$ by conditional maximum likelihood:
\[
\max_{\boldsymbol{\phi}}\ \sum_{x}\ \frac{1}{S}\sum_{s=1}^{S}
\log q_{\boldsymbol{\phi}}\!\big(\mathbf{z}^{(s)}\mid \mathbf{h}(x)\big).
\]

\subsection{Extracting Uncertainty Signals}
\label{sec:distribution-uses}

At inference time, SSD produces a prompt-conditioned semantic density $q_\phi(\mathbf{z}\mid\mathbf{h})$ over answer embeddings, which we treat as a general-purpose uncertainty object. In this work, we focus on two primary operations derived from this object: prompt-level dispersion and answer-level likelihood. We also highlight that the explicit density enables additional distributional operations such as sampling.

\textbf{(i) Prompt evaluation.}
We use the entropy of the predicted semantic distribution $q_\phi(\cdot\mid\mathbf{h})$ to provide a prompt-level uncertainty estimate in a single forward pass.

\textbf{(ii) Answer evaluation.}
Given any candidate response $\mathbf{y}'$, we embed it as $\mathbf{z}'=\Phi(\mathbf{y}')$ and score its compatibility with the prompt via the posterior log-likelihood $\log q_\phi(\mathbf{z}'\mid\mathbf{h})$. We may treat answers with low likelihood as out-of-domain relative to the prompt context.

\textbf{(iii) Latent sampling.}
Because $q_\phi(\mathbf{z}\mid\mathbf{h})$ is an explicit mixture distribution, we can compute summary statistics such as the mixture mean
$\mathbb{E}_{q_\phi}[\mathbf{z}]=\sum_k \pi_k(\mathbf{h})\,\boldsymbol{\mu}_k(\mathbf{h})$
and, when desired, draw latent samples $\tilde{\mathbf{z}}\sim q_\phi(\cdot\mid\mathbf{h})$.

We investigate operations (i) and (ii) in Sections \ref{sec:results-prompt-level} and \ref{sec:results-answer-level}, respectively. Operation (iii) illustrates a structural consequence of modelling the full predictive density and is discussed in Appendix~\ref{sec:results-sampling}.

\subsubsection{Quantifying Dispersion}
We measure dispersion using the order-2 Rényi entropy \citep{renyi1961measures} of the predictive density $q_\phi(\cdot\mid\mathbf{h})$:
\[
\mathsf{H}_2\!\big(q_\phi(\cdot\mid \mathbf{h})\big)
= -\log \int q_\phi(\mathbf{z}\mid \mathbf{h})^2 \, d\mathbf{z}.
\]
Intuitively, the integral $\int q(\mathbf{z})^2 d\mathbf{z}$ is the \emph{collision probability} of the density: it is larger when mass is concentrated, and smaller when the distribution is spread out \citep{principe2010information}. Higher $\mathsf{H}_2$ corresponds to greater semantic dispersion.

We use Rényi-2 because it yields an exact closed form for Gaussian mixtures \citep{wang2009closed}, enabling a sampling-free, analytic dispersion score at inference time. By contrast, the Shannon differential entropy \citep{shannon1948mathematical}
does not, in general, admit a closed-form expression for Gaussian mixture models; it is typically approximated using Monte Carlo estimation or via analytic bounds \citep{huber2008entropy, dahlke2023closedform}. Using Rényi-2 therefore preserves SSD's goal of avoiding additional sampling or expensive numerical estimation at deployment. We provide the closed form expression and derivation in Appendix~\ref{app:theory}.

\subsection{Extension to Generic Sequence Domains}
\label{sec:methods_extension}

Although we focus on autoregressive language models in this work, the SSD construction is agnostic to modality. In language tasks, $\mathbf{z} = \Phi(y_{1:L})$ captures the semantic content of a generated answer. More generally, $\mathbf{z}$ may represent any task-relevant latent characterization of a predicted sequence $y_{1:L}$, rather than its individual tokens or events.

Under stochastic generation, a sequence model induces a predictive distribution over such latent representations. SSD distils this sampled distribution into a lightweight conditional density $q_\phi(\mathbf{z} \mid \mathbf{h})$, where $\mathbf{h}$ encodes the input sequence. The resulting density provides a \textit{sampling-free approximation to the model's belief over high-level outcomes}. We provide further discussion of potential application domains in Section \ref{sec:future-work}.

\section{Related Work}

\paragraph{Uncertainty estimation in language models.}
\citet{xiong2024efficient} categorize UQ methods for LLMs into three broad categories: single-sample, multi-sample, and probing-based. Single-sample methods are training-free and include output perplexity, negative sequence probability, and mean token entropy \citep{malinin2021uncertainty, aichberger2026rethinking}. 

Probing-based methods similarly use a single sample at inference time but train a predictive model on internal LLM activations to predict an uncertainty-relevant quantity, such as the probability of correctness \citep{kadavath2022languagemodelsmostlyknow, liu2024uncertaintyestimationquantificationllms}, semantic dispersion \citep{kossen2024semanticentropyprobesrobust}, or token-level density \citep{vazhentsev-etal-2025-token}. Related supervised approaches fine-tune the model itself
\citep{kapoor2024large} or learn a scoring function over token
probabilities \citep{yaldiz-etal-2025-design}.

Multi-sample methods often rely on notions of semantic dispersion, either explicitly, such as \textit{Semantic Entropy} \citep{farquhar_detecting_2024}, \textit{Number of Clusters} \citep{lin2024generating}, and \textit{EigenScore} \citep{chen2024inside}, or implicitly, as with \textit{P(true)}~\citep{kadavath2022languagemodelsmostlyknow}. Further consistency-based variants aggregate agreement across sampled generations or perturbed inputs \citep{gao-etal-2024-spuq, bhattacharjya-etal-2025-simba}, a premise formalized by the consistency hypothesis \citep{xiao2025the}. We emphasize that these approaches typically model a \textit{proxy} for language model uncertainty, geared for application in a particular task. There is limited consensus on how to decompose classical notions of aleatoric and epistemic uncertainty in LLMs, and the resulting uncertainty signals are often only modestly correlated \citep{hou2024decomposing, xiong2024efficient}.

\paragraph{Distillation in language models.}
Distillation typically refers to training smaller language models on the \textit{logit} or \textit{token} outputs of more capable models, with notable success in reasoning domains \citep{guo2025deepseek}. Such approaches transfer predictive behaviour but do not explicitly target calibrated uncertainty or structured output distributions beyond the token level. In contrast, we focus on \textit{sequence-level} distributions of model outputs and investigate whether these can be elicited without autoregressive sampling. \citet{piskorz2026eliciting} study a related but narrower phenomenon concerning language model predicted distributions over numerical targets, whereas we consider broader semantic targets.  

\paragraph{Knowledge distillation for uncertainty estimation.}
A related line of work seeks to distil ensemble-based uncertainty estimates into a single model~\citep{Malinin2020Ensemble, fathullah2022self, landgraf2024dudes}. Ensembles are widely regarded as producing high-quality uncertainty estimates, but incur substantial computational cost at inference time. Recent approaches therefore train a student model to approximate ensemble predictive distributions or uncertainty signals, aiming to recover ensemble-quality uncertainty in a single forward pass, typically by matching logits, predictive variances, or other distributional summaries~\citep{fathullah2023logit, park2025knowledge}.

In contrast, our setting does not assume access to an external ensemble and does not aim to replicate variance estimates or token-level predictive distributions. Instead, we distil a language model's \emph{semantic sequence-level output distribution} induced by stochastic sampling, learning a prompt-conditioned density over answer embeddings. The most closely related approaches in the LLM literature are probing-based uncertainty estimators~\citep{kadavath2022languagemodelsmostlyknow, kossen2024semanticentropyprobesrobust}, which however compress uncertainty into a single scalar statistic. By contrast, we predict a \emph{full sequence-level distribution} in a single forward pass, enabling downstream operations beyond scalar risk scoring, including likelihood-based post-generation verification and multiple-choice answer selection. Rather than estimating a task-specific uncertainty heuristic, we approximate the underlying semantic output distribution itself.

\begin{figure*}[t]
    \centering
    \includegraphics[trim={0cm 14.30cm 1cm 0cm}, clip,width=\linewidth]{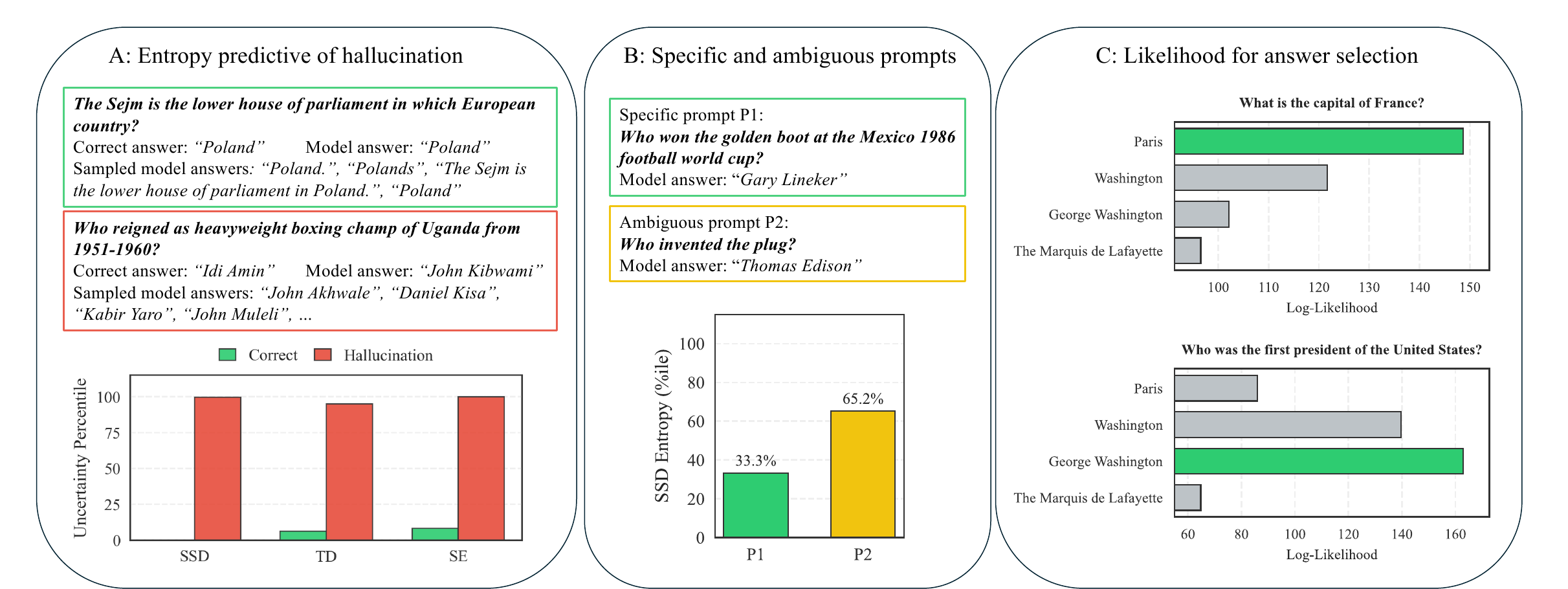}
   \caption{\textbf{Qualitative analysis of the SSD density.}
\textbf{(A)} A correct answer yields minimal SSD entropy (0.1\textsuperscript{st} percentile), while a hallucination yields high uncertainty (99.7\textsuperscript{th}), aligning with sampling-based Teacher Dispersion (TD) and Semantic Entropy (SE). Percentiles denote rank within the TriviaQA test set ($N{=}1{,}000$).
\textbf{(B)} A precise prompt (P1) yields significantly lower entropy than an underspecified one (P2).
\textbf{(C)} The SSD likelihood enables context-aware scoring: given identical candidate sets, the student assigns the highest likelihood to the semantically appropriate answer for each prompt.}
\label{fig:qualitative_analysis}
\end{figure*}

\section{Experiments}
\label{sec:experiments}

We evaluate SSD across multiple LLM families in order to answer two core questions. 
First, does SSD faithfully distil sampling-based semantic dispersion into a single-pass model? 
Second, what additional capabilities arise from modelling a full sequence-level density rather than a scalar uncertainty proxy? To address the first question, we study \emph{prompt-level} hallucination prediction, treating the student's entropy as a \emph{pre-generation} risk score and comparing it against sampling-based and probing baselines. To address the second, we evaluate \emph{answer-level} likelihood scores, demonstrating how the predicted density supports \emph{post-generation} verification, including out-of-domain answer detection and multiple-choice answer selection. Figure~\ref{fig:qualitative_analysis} illustrates examples of each operation type.

\subsection{Models and Dataset Generation}
\label{sec:experiments_models_datasets}

We evaluate SSD in the context of short-form question answering, using the TriviaQA~\citep{joshi-etal-2017-triviaqa} and MMLU~\citep{hendrycks2021measuring} datasets. We construct training sets of $N_{train}=4{,}000$ prompts and report results on held-out test sets of $N_{test}=1{,}000$ prompts.

MMLU is formatted as a multiple-choice benchmark, as opposed to the open-ended TriviaQA. To properly assess semantic dispersion and enable comparison across datasets, we use a filtered subset of MMLU containing only open-ended questions \citep{myrzakhan2024openllmleaderboardmultichoiceopenstylequestions}. 

We select these datasets because they represent two distinct task types: \textbf{knowledge-seeking} (TriviaQA) and \textbf{reasoning-heavy} (MMLU). Prior work has found semantic dispersion to be a useful proxy for language model uncertainty in the former task type and less so in the latter \citep{xiong2024efficient}; as such MMLU provides a more challenging setting for evaluating SSD, although in practice it reflects a mixture of both task types.

We evaluate across five model families: \textbf{Mistral} (Ministral-8B-Instruct), \textbf{Llama-3} (3.1-8B-Instruct, 3.2-3B-Instruct), \textbf{Qwen} (Qwen3-4B-Instruct-2507, Qwen3-8B), \textbf{Gemma} (Gemma-3-4B) and \textbf{SmolLM} (SmolLM3-3B)~\citep{grattafiori2024llama, mistralai_ministraux_2024, yang2025qwen3technicalreport, gemmateam2025gemma3technicalreport, bakouch2025smollm3}. These models span multiple architectural families, parameter scales (3B-8B), and training pipelines, allowing us to assess the robustness and generality of SSD across diverse backbone designs rather than tailoring it to a single model.

For all experiments, we use Qwen3-8B as the oracle `judge' model to determine the correctness of generated answers given a reference answer~\citep{zheng2023judge}. For each prompt and model, we generate $S=32$ stochastic samples at temperature $T=1$. We additionally sample a single `default' answer at $T=0.1$, which is provided to the judge model to obtain the hallucination label.

For answer embeddings, we use EmbeddingGemma~\citep{vera2025embeddinggemmapowerfullightweighttext}, with the embedding truncated to a dimension of $d_z=128$. For hallucination prediction, we further reduce dimensionality using PCA to $d_z'=16$, following ablation studies detailed in Appendix~\ref{sec:ablations}. For likelihood-based OOD verification, we set $d_z'=32$.

The student MDNs themselves are lightweight: across the configurations selected in Appendix~\ref{sec:model-selection}, total parameter counts range from approximately 600K to 10M depending on hidden width, depth, and the base model's embedding dimension - orders of
magnitude smaller than the 3B-8B parameter base LLMs.

\subsection{Prompt-Level Hallucination Prediction}
\label{sec:results-prompt-level}

\begin{table*}[t]
    \centering
    \caption{\textbf{Hallucination prediction trade-offs.} AUROC is reported per dataset as a macro-average (mean $\pm$ std) across models. Latency summarizes inference-time requirements (single-pass vs. $S$ sampled generations, with entailment-based SE additionally requiring $S^2$ natural language inference (NLI) comparisons). Target cost denotes the computation required to construct supervision targets during training; $C_{CL}$ is the cost of acquiring correctness labels. Only SSD produces an explicit predictive density, enabling post-generation likelihood scoring and latent sampling in addition to pre-generation uncertainty.}
    \label{tab:method-attribs}
	\resizebox{0.85\linewidth}{!}{%
    \begin{tabular}{lccccc}
\toprule
Method & TriviaQA $\uparrow$ & MMLU $\uparrow$ & Latency $\downarrow$ & Target cost $\downarrow$ & Predictive density\\
\midrule
SE & \textbf{0.804 $\pm$ 0.035} & 0.663 $\pm$ 0.040 & $S$ samples + $S^2$ NLI & -- & \noPosterior \\
SEP & 0.759 $\pm$ 0.026 & 0.628 $\pm$ 0.029 & \textbf{Single-pass} & $S$ samples + $S^2$ NLI & \noPosterior \\
PCP & 0.771 $\pm$ 0.020 & \textbf{0.679 $\pm$ 0.019} & \textbf{Single-pass} & $C_{\text{CL}}$ & \noPosterior \\
TD & 0.699 $\pm$ 0.027 & 0.647 $\pm$ 0.023 & $S$ samples & -- & \noPosterior \\
SSD & 0.708 $\pm$ 0.024 & 0.623 $\pm$ 0.025 & \textbf{Single-pass} & $S$ samples + $S\,\Phi$ & \yesPosterior \\
\bottomrule
\end{tabular}
    }
\end{table*}

\begin{table}[t]
    \centering
   \caption{\textbf{Hallucination prediction, comparison with teacher dispersion.}
   TD computes dispersion from $S=32$ sampled semantic embeddings at inference time, while SSD predicts a prompt-conditioned semantic density and scores dispersion analytically.
   We report hallucination prediction AUROC/AUPRC in \% as the mean over 1{,}000 bootstrap resamples of the test set, with standard deviation as subscript; bold indicates the better of TD vs.\ SSD per metric and dataset.}
    \label{tab:ed-vs-ssd}
    \centering
	\resizebox{0.875\linewidth}{!}{%
    \begin{tabular}{lcccc}
\toprule
 & \multicolumn{2}{c}{AUROC} & \multicolumn{2}{c}{AUPRC} \\
\cmidrule(lr){2-3} \cmidrule(lr){4-5}
Model & TD & SSD & TD & SSD \\
\midrule
\multicolumn{5}{l}{\textbf{TriviaQA}} \\
\midrule
Qwen3 8B & \textbf{68.3\textsubscript{1.6}} & 67.7\textsubscript{1.8} & 52.4\textsubscript{2.6} & \textbf{56.9\textsubscript{2.8}} \\
Qwen3 4B & \textbf{73.2\textsubscript{1.6}} & 69.9\textsubscript{1.6} & 59.7\textsubscript{2.4} & \textbf{61.9\textsubscript{2.5}} \\
Llama 3.1 8B & \textbf{69.6\textsubscript{1.7}} & 68.7\textsubscript{1.9} & 44.4\textsubscript{2.7} & \textbf{51.9\textsubscript{2.9}} \\
Llama 3.2 3B & 66.6\textsubscript{1.7} & \textbf{73.7\textsubscript{1.6}} & 55.8\textsubscript{2.6} & \textbf{65.5\textsubscript{2.5}} \\
Ministral 8B & 74.3\textsubscript{1.7} & \textbf{74.9\textsubscript{1.6}} & 57.9\textsubscript{2.8} & \textbf{59.9\textsubscript{2.8}} \\
SmolLM3 3B & 69.9\textsubscript{1.6} & \textbf{71.1\textsubscript{1.6}} & 57.1\textsubscript{2.5} & \textbf{62.4\textsubscript{2.5}} \\
Gemma 3 4B & 67.6\textsubscript{1.6} & \textbf{70.0\textsubscript{1.7}} & 63.1\textsubscript{2.3} & \textbf{66.1\textsubscript{2.5}} \\
\midrule
\multicolumn{5}{l}{\textbf{MMLU}} \\
\midrule
Qwen3 8B & \textbf{63.5\textsubscript{1.7}} & 57.1\textsubscript{1.8} & \textbf{61.9\textsubscript{2.3}} & 55.9\textsubscript{2.3} \\
Qwen3 4B & 63.1\textsubscript{1.8} & \textbf{64.9\textsubscript{1.7}} & 58.5\textsubscript{2.4} & \textbf{59.4\textsubscript{2.5}} \\
Llama 3.1 8B & \textbf{69.0\textsubscript{1.6}} & 63.8\textsubscript{1.7} & \textbf{70.9\textsubscript{2.1}} & 62.7\textsubscript{2.4} \\
Llama 3.2 3B & \textbf{67.1\textsubscript{1.8}} & 63.8\textsubscript{1.7} & 74.0\textsubscript{2.0} & \textbf{74.5\textsubscript{1.9}} \\
Ministral 8B & \textbf{65.1\textsubscript{1.8}} & 63.6\textsubscript{1.8} & 71.4\textsubscript{2.1} & \textbf{71.7\textsubscript{2.1}} \\
SmolLM3 3B & \textbf{62.3\textsubscript{1.8}} & 62.0\textsubscript{1.8} & \textbf{64.6\textsubscript{2.2}} & 63.0\textsubscript{2.3} \\
Gemma 3 4B & \textbf{62.8\textsubscript{1.7}} & 60.7\textsubscript{1.8} & \textbf{71.1\textsubscript{2.0}} & 69.2\textsubscript{2.1} \\
\bottomrule
\end{tabular}
    }
\end{table}

We first evaluate whether the distilled semantic density provides a reliable \emph{pre-generation} hallucination signal. Specifically, we treat the entropy of the predicted distribution as a prompt-level risk score: higher entropy corresponds to greater semantic dispersion and increased hallucination risk. We report AUROC and AUPRC against correctness labels determined using an LLM-as-judge framework.

\paragraph{SSD compares favourably to teacher dispersion.}
We compare the student's dispersion score to the standard deviation of the teacher's $S=32$ semantic embeddings, which we term \textit{teacher dispersion} (TD). Table~\ref{tab:ed-vs-ssd} reports results across models and datasets. On the TriviaQA dataset, despite requiring no inference-time sampling, SSD outperforms TD on four of the seven models investigated in terms of AUROC and on all seven in AUPRC. We attribute this in part to a smoothing effect: TD is a finite-sample statistic and assigns zero uncertainty to prompts where the model consistently produces the same answer, even when that answer is incorrect. By contrast, SSD learns a continuous mapping from prompt representations to semantic uncertainty, enabling differentiated risk estimates in such degenerate cases, which likely contributes to its stronger AUPRC.

On the MMLU dataset, SSD is a relatively weaker performer compared to the sampled teacher signal, which we attribute to the task being less suited for the semantic dispersion uncertainty proxy. SSD still surpasses TD in three out of seven models from an AUPRC perspective. 

Overall, these results demonstrate that SSD successfully distils sampling-based semantic dispersion into a single-pass density estimator. Despite requiring no inference-time sampling, the student recovers much of the teacher's uncertainty signal, and in some cases improves upon the finite-sample baseline. The distilled signal is also not narrowly tied to its training distribution: students evaluated on the dataset they were not trained on lose only 0.04-0.05 mean AUROC relative to in-distribution students
(Appendix~\ref{app:cross_dataset}).

\paragraph{Comparison to additional baselines.}
We compare SSD against three widely used hallucination prediction methods: (1) \textbf{Probability of Correctness Probe (PCP)}, a logistic regression classifier trained on the prompt representation to predict correctness. Originally proposed as `P(I know)' by \citet{kadavath2022languagemodelsmostlyknow}, this method requires externally provided correctness labels. (2) \textbf{Semantic Entropy (SE)}, a sampling-based approach that measures semantic dispersion by clustering $S$ completions using a natural language inference (NLI) model to assess mutual entailment~\citep{farquhar_detecting_2024}. (3) \textbf{Semantic Entropy Probe (SEP)}, a lightweight probe with the same architecture as PCP, trained to regress a binarized SE target, and inheriting SE's sampling and NLI-based target construction cost \citep{kossen2024semanticentropyprobesrobust}.

To provide a fuller comparison of methods, we report hallucination prediction performance alongside method characteristics: inference latency, training supervision target construction cost, and whether a method supports post-generation signals in addition to pre-generation uncertainty. Table~\ref{tab:method-attribs} summarizes these trade-offs and reports macro-average AUROC across model families; full per-model AUROC/AUPRC tables are provided in Appendix~\ref{app:full-hd-results}. We provide further analysis of the computational complexity of each method in Appendix~\ref{app:complexity}.

Overall, SE achieves the highest AUROC on TriviaQA, whilst PCP is the strongest performer on MMLU, consistent with prior work~\citep{xiong2024efficient}. While SE and PCP achieve the best performance in their respective regimes, these methods either incur significant sampling cost (SE) or reduce uncertainty to a supervised scalar classification problem (PCP). SSD occupies a distinct regime: it matches probe-level inference latency, avoids entailment-based target construction during training, and preserves an explicit predictive density. Although its pre-generation hallucination prediction performance is slightly weaker than specialized probes, it uniquely provides post-generation likelihood signals derived from the same model. 

We further observe that the SSD performance is model-dependent. On TriviaQA, for example, it achieves an AUROC of 74.9 for Ministral 8B but only 67.7 for Qwen3 8B (Table~\ref{tab:main_results}). 
Analysis in Appendix~\ref{app:dist-fidelity} indicates that this variation is driven primarily by the \textit{fidelity} of the distillation process: for some models, it is particularly difficult to predict the semantic distribution from the prompt representation. We discuss these findings further in Section~\ref{sec:future-work}.

\subsection{Answer-Level Response Evaluation}
\label{sec:results-answer-level}

Having established that SSD recovers sampling-based dispersion signals, we next evaluate the additional capabilities enabled by modelling an explicit predictive density. Unlike scalar uncertainty scores, the density allows us to assess the quality of \emph{specific} candidate answers by measuring their likelihood under the model's predicted semantic distribution. We examine this utility in two settings: out-of-domain answer detection (binary classification) and multiple-choice question answering (multi-class selection).

\subsubsection{Out-of-domain Answer Detection}
\label{sec:ood}
We first  evaluate the student's ability to detect mismatched prompt-answer pairs. We frame this as an out-of-domain detection task: for each test prompt $x_i$, we pair its representation $\mathbf{h}_i$ with (i) the model's default answer $y_i$ and (ii) a random answer $y_j$ drawn from a different prompt. Each candidate answer is embedded as $\mathbf{z}=\Phi(y)$ and scored using the student's log-likelihood $\log q_\phi(\mathbf{z}\mid\mathbf{h})$. We report AUROC for distinguishing matched from mismatched pairs.

As shown in Table~\ref{tab:utility_likelihood}, the SSD log-likelihood is a strong context-verification signal, achieving AUROCs above 0.95 for most base models on both TriviaQA and MMLU. Practically, this could enable SSD to act as a content `guard'~\citep{inan2023llamaguardllmbasedinputoutput}, for example detecting cache poisoning \citep{chenagentpoison2024} or filtering retrieved content in retrieval-augmented generation \citep{lewisrag2020}. This mismatch-detection capability also generalizes across distribution shift: students evaluated on the dataset they were not trained on retain AUROCs of 0.75--0.84 (Appendix~\ref{app:cross_dataset}).

We also evaluate the student likelihood as a \textit{post-generation} hallucination signal, motivated by prior work showing that probability density in semantic space is predictive of correctness~\citep{qiu2024semanticdensity}. While informative, our likelihood signal is weaker than the pre-generation entropy, with AUROC scores of approximately 0.6. Unlike sampling-based density methods, which construct the semantic distribution from multiple generations at inference time, SSD predicts the density from the prompt representation alone. The weaker performance reflects the limits of this single-pass approximation to a sampling-derived density. 

We therefore do not advocate the likelihood as a stand-alone hallucination predictor. We report these results to establish that the signal is informative (AUROC above 0.5) and complementary in principle to the pre-generation entropy; the primary post-generation use case remains the mismatch detection task above.

\begin{table}[t]
    \centering
    \caption{\textbf{Out-of-domain answer detection.} We report AUROC for two binary classification tasks using the student's log-likelihood score. \textbf{Mismatch} measures the ability to distinguish the model's answer to the prompt from a random answer from the dataset. \textbf{Hallucination} measures the ability to distinguish correct model-generated answers from incorrect ones.
    Values report mean AUROC with bootstrap standard deviation over 1{,}000 resamples.}
    \label{tab:utility_likelihood}
	\resizebox{0.85\linewidth}{!}{%
    \begin{tabular}{lcc}
\toprule
Model & Mismatch & Hallucination\\
\midrule
\multicolumn{3}{l}{\textbf{TriviaQA}}\\
\midrule
Qwen3 8B & 0.984 $\pm$ 0.002 & 0.571 $\pm$ 0.019 \\
Qwen3 4B & 0.984 $\pm$ 0.002 & 0.557 $\pm$ 0.018 \\
Llama 3.1 8B & 0.989 $\pm$ 0.002 & 0.579 $\pm$ 0.019 \\
Llama 3.2 3B & 0.980 $\pm$ 0.003 & 0.613 $\pm$ 0.017 \\
Ministral 8B & 0.945 $\pm$ 0.005 & 0.616 $\pm$ 0.017 \\
SmolLM3 3B & 0.988 $\pm$ 0.002 & 0.562 $\pm$ 0.018 \\
Gemma 3 4B & 0.895 $\pm$ 0.007 & 0.578 $\pm$ 0.018 \\
\midrule
\multicolumn{3}{l}{\textbf{MMLU}}\\
\midrule
Qwen3 8B & 0.975 $\pm$ 0.003 & 0.526 $\pm$ 0.018 \\
Qwen3 4B & 0.960 $\pm$ 0.004 & 0.566 $\pm$ 0.019 \\
Llama 3.1 8B & 0.981 $\pm$ 0.003 & 0.576 $\pm$ 0.018 \\
Llama 3.2 3B & 0.942 $\pm$ 0.005 & 0.519 $\pm$ 0.019 \\
Ministral 8B & 0.966 $\pm$ 0.004 & 0.529 $\pm$ 0.019 \\
SmolLM3 3B & 0.985 $\pm$ 0.003 & 0.528 $\pm$ 0.018 \\
Gemma 3 4B & 0.923 $\pm$ 0.006 & 0.548 $\pm$ 0.019 \\
\bottomrule
\end{tabular}
    }
\end{table}

\subsubsection{Multiple-choice Answer Selection}
\label{sec:results-mcq}

We next evaluate whether the student's predicted density can be used to identify the correct  answer in a multiple-choice setting. We utilize the open-ended subset of MMLU provided by \citet{myrzakhan2024openllmleaderboardmultichoiceopenstylequestions}, which filters for questions that are self-contained and do not require the context of multiple-choice options to answer. 

The student is trained using only open-ended completions to prompts in the training set. For the test set, we take the four provided options ($A, B, C, D$) for each question, embed them using $\Phi$, and rank them according to their log-likelihood $\log q_\phi(\mathbf{z}'\mid\mathbf{h})$ under the student mixture.

Whereas the mismatched answers in Section \ref{sec:ood} are semantically distant from the prompt by construction, the four MMLU options are deliberately close in semantic space, so this experiment poses a more demanding test: can the density discriminate fine-grained semantic structure within the plausible answer region?

Table~\ref{tab:ssd_choice_selection} reports the accuracy of selecting the highest-likelihood option (\textbf{Top-1}) alongside the base model's open-ended answer accuracy as assessed by an LLM-as-judge. SSD consistently exceeds the 0.25 random baseline, and for our best performing models (Llama 3.2 and Ministral) approaches the base model's open-ended accuracy. 

The open-ended accuracy of the base model provides a rough estimate of its dataset-specific knowledge. Despite being trained only on sampled open-ended completions and without multiple-choice supervision, the student approaches this level of performance when provided with the options, suggesting that the distilled semantic density captures meaningful structure in the model's answer distribution. We emphasize that SSD is not proposed as a substitute for directly querying the model; the purpose of the experiment is to validate the predicted density as a usable belief state.

\begin{table}[t]
    \centering
    \caption{\textbf{Multiple-choice answer selection.} We evaluate the ability of the SSD student -  trained solely on open-ended generations - to identify the correct answer in a multiple-choice setting using an open-ended subset of MMLU. We report \textbf{Top-1} accuracy, where the option with the highest predicted likelihood is selected (random baseline $= 0.25$). \textbf{Open-Ended} accuracy denotes the base model's accuracy when generating answers directly, as determined by an LLM-as-judge.}
    \label{tab:ssd_choice_selection}
	\resizebox{0.85\linewidth}{!}{%
    \begin{tabular}{lcc}
\toprule
Model & SSD Top-1 & Open-Ended\\
\midrule
Qwen3 8B & 0.327 $\pm$ 0.015 & 0.515 $\pm$ 0.015 \\
Qwen3 4B & 0.312 $\pm$ 0.014 & 0.545 $\pm$ 0.016 \\
Llama 3.1 8B & 0.339 $\pm$ 0.015 & 0.483 $\pm$ 0.015 \\
Llama 3.2 3B & 0.346 $\pm$ 0.015 & 0.358 $\pm$ 0.015 \\
Ministral 8B & 0.363 $\pm$ 0.015 & 0.396 $\pm$ 0.015 \\
SmolLM3 3B & 0.332 $\pm$ 0.015 & 0.466 $\pm$ 0.016 \\
Gemma 3 4B & 0.292 $\pm$ 0.014 & 0.399 $\pm$ 0.015 \\
\bottomrule
\end{tabular}
    }
\end{table}

\paragraph{Summary of experimental findings.}
Taken together, the prompt-level and answer-level evaluations demonstrate that SSD's value lies not in a particular scalar uncertainty score, but in exposing an explicit semantic belief state from which multiple reliability signals can be derived. By modelling a prompt-conditioned distribution rather than a scalar proxy, SSD unifies pre-generation risk estimation and post-generation verification within a single, low-latency framework. The performance variation observed across base LLMs points to differences in how semantic uncertainty is encoded in internal activations. 

We therefore position SSD as an appropriate UQ tool when inference latency precludes sampling-based methods, when the labelled correctness data required by supervised probes is unavailable or expensive, or when post-generation answer-level operations are needed alongside pre-generation risk estimates. 
Practically, we would recommend computing the distillation fidelity (discussed in Section \ref{sec:future-work} and Appendix \ref{app:dist-fidelity}) on a held-out validation set before using SSD in deployment. A student that fails to track teacher dispersion at training time should not be relied upon for risk estimation at inference.


\section{Discussion \& Future Work}
\label{sec:future-work}

The experimental results highlight both the strengths and current limitations of SSD. While modelling a full semantic density enables richer post-generation capabilities than scalar probes, its effectiveness depends on how faithfully the student can reconstruct the teacher's sampled distribution from prompt representations alone. We now discuss these limitations and outline directions for strengthening the distributional framework.

\paragraph{Improving distillation fidelity.} 
Our results in Appendix~\ref{app:dist-fidelity} show a strong correlation between distillation fidelity and hallucination prediction performance, indicating that SSD's effectiveness depends on how accurately the student reconstructs the teacher's prompt-induced uncertainty ordering. Promising directions to improve this fidelity include scaling the distillation dataset size, and distilling from stronger teachers (for example sampling more responses per question, or using higher-quality semantic targets). 

We believe the most significant gains could be found by enriching the student input beyond a single-token/single-layer representation, something we explore in Appendix~\ref{app:richer_reps}. Using only simple enrichment approaches, we find modest and model-dependent gains in SSD performance, motivating more principled investigation of optimal probe design \citep{kramar2026buildingproductionreadyprobesgemini}. Subsequent findings may clarify whether observed performance gaps between models stem from limited semantic information in the prompt representation or insufficient supervision for learning high-dimensional densities.


\paragraph{Combining pre- and post-generation signals.}
SSD exposes complementary reliability signals from a single predicted density: pre-generation dispersion forecasts risk, while post-generation likelihood verifies prompt–answer compatibility. A natural next step is to integrate these signals into unified decision policies for LLM-based systems, for example abstaining or retrieving evidence under high dispersion, and accepting or rejecting candidate actions based on likelihood thresholds.

\textbf{Diffusion models.} Although our experiments focus on autoregressive models, SSD may be particularly well suited to diffusion-based language models~\citep{sahoodiffusion2024}. In autoregressive decoding, the model's belief over the final answer is implicit in a long sequence of conditional token distributions. In masked diffusion models, by contrast, the model produces predictions over the full output sequence at every denoising step, so sequence-level information is more concentrated in early internal states. The mapping from prompt representation to final output distribution that SSD must learn may therefore be more direct, potentially reducing the gap between the model's internal belief state and the semantic distribution induced by sampling.

\paragraph{Domain-specific uncertainty estimation.}
As described in Section~\ref{sec:methods_extension}, SSD extends beyond language to sequence models whose outputs admit meaningful latent representations. We see particular potential utility for this technique in electronic health record (EHR) foundation models, which tokenize longitudinal patient histories and generate future clinical event sequences~\citep{renc2024zero}. These models typically predict discrete events, but clinical utility often depends on a higher-level characterization of the patient's future state, such as overall disease burden or risk profile~\citep{steinberg2024motor, zeng2025trajsurvlearningcontinuouslatent}. 

In this setting, the embedding model could correspond to a transformer-based EHR encoder such as BEHRT~\citep{li2020behrt}, summarizing a predicted trajectory into a clinically informative latent health representation. Sampling from $q_\phi(\mathbf{z} \mid \mathbf{h})$ could then yield low-latency forecasts of plausible future patient states, while the entropy of the density would quantify uncertainty over long-term outcomes. Similar structure arises in time-series foundation models~\citep{faw2025incontext}, where future trajectories may be summarized by latent representations capturing regime, trend, or seasonal structure~\citep{wang2022learning}.

\paragraph{LLM-as-judge evaluation.} Our correctness labels derive from a single judge model (Qwen3-8B). LLM-as-judge has been found to be the least length-biased among common correctness functions \citep{santilli-etal-2025-revisiting}, supporting this protocol choice, but judge-induced bias cannot be excluded, and marginalizing over multiple judge variants could offer more reliable labels in future work \citep{ielanskyi2026addressing}.

\paragraph{Scaling to larger models.} SSD requires $S$ stochastic completions per training prompt, a one-time offline cost that grows with model scale. In practice, such datasets are routinely produced during RL-based post-training, where multiple rollouts per prompt are generated by algorithms such as PPO \citep{schulman2017proximalpolicyoptimizationalgorithms} and GRPO \citep{shao2024deepseekmathpushinglimitsmathematical}. Where such pipelines exist, the marginal cost of SSD implementation is small, even for very large models.

\section{Conclusion}

We introduced Semantic Self-Distillation (SSD), a framework for distilling a language model's sampled semantic answer distribution into a lightweight, prompt-conditioned predictive density. Rather than reducing uncertainty to a scalar score, SSD models the geometry of the semantic output space itself, enabling analytic dispersion estimates and likelihood-based evaluation in a single forward pass.

Empirically, SSD performs competitively with the teacher's sampled semantic dispersion across several model families, while avoiding the inference-time overhead of generating multiple completions. Although it does not dominate specialized scalar probes on any single task, SSD uniquely combines probe-level latency, freedom from correctness-label and entailment-based supervision, and an explicit predictive density within a single estimator. Crucially, modelling the full density enables capabilities unavailable to scalar methods, including answer-level reliability scoring and distribution-based selection.

More broadly, SSD demonstrates that uncertainty in complex generative models can be distilled as an explicit density over latent outcome representations. By shifting computational cost from inference to training, SSD provides a practical mechanism for integrating principled, distribution-level uncertainty into latency-critical language systems and other sequence domains.



\begin{acknowledgements} 
    DAC was funded by an NIHR Research Professorship; a Royal Academy of Engineering Research Chair; and the InnoHK Hong Kong Centre for Cerebro-cardiovascular Engineering (COCHE); and was supported by the National Institute for Health Research (NIHR) Oxford Biomedical Research Centre (BRC) and the Pandemic Sciences Institute at the University of Oxford. EP was funded by an NIHR Research Studentship. SW was supported by the Rhodes Scholarship.

\end{acknowledgements}

\bibliography{uai2026-template}

\renewcommand{\thefigure}{A\arabic{figure}}
\setcounter{figure}{0}
\renewcommand{\thetable}{A\arabic{table}}
\setcounter{table}{0}
\renewcommand{\theequation}{A\arabic{equation}}
\setcounter{equation}{0}

\newpage

\onecolumn

\title{Semantic Self-Distillation for Language Model Uncertainty\\(Appendix)}
\maketitle

\appendix

\section{Derivations}
\label{app:theory}
\paragraph{Closed-form computation of Rényi–2 entropy.} For a Gaussian mixture, the quadratic functional $\int q(\mathbf{z})^2 d\mathbf{z}$ decomposes into pairwise overlaps between mixture components:
\[
\int q_\phi(\mathbf{z}\mid\mathbf{h})^2\, d\mathbf{z}
=
\sum_{i=1}^{K}\sum_{j=1}^{K}\pi_i(\mathbf{h})\,\pi_j(\mathbf{h})
\int \mathcal{N}(\mathbf{z};\boldsymbol{\mu}_i,\boldsymbol{\Sigma}_i)\,
     \mathcal{N}(\mathbf{z};\boldsymbol{\mu}_j,\boldsymbol{\Sigma}_j)\, d\mathbf{z}.
\]
Using the standard Gaussian product identity,
\[
\int \mathcal{N}(\mathbf{z};\boldsymbol{\mu}_i,\boldsymbol{\Sigma}_i)\,
     \mathcal{N}(\mathbf{z};\boldsymbol{\mu}_j,\boldsymbol{\Sigma}_j)\, d\mathbf{z}
=
\mathcal{N}\!\big(\boldsymbol{\mu}_i;\boldsymbol{\mu}_j,\boldsymbol{\Sigma}_i+\boldsymbol{\Sigma}_j\big),
\]
we obtain:
\[
\mathsf{H}_2\!\big(q_\phi(\cdot\mid \mathbf{h})\big)
=
-\log\!\left(
\sum_{i=1}^{K}\sum_{j=1}^{K}
\pi_i(\mathbf{h})\,\pi_j(\mathbf{h})\,
\mathcal{N}\!\big(\boldsymbol{\mu}_i(\mathbf{h});\boldsymbol{\mu}_j(\mathbf{h}),
\boldsymbol{\Sigma}_i(\mathbf{h})+\boldsymbol{\Sigma}_j(\mathbf{h})\big)
\right).
\]
Equivalently, defining the collision matrix $\mathbf{K}(\mathbf{h})\in\mathbb{R}^{K\times K}$ with entries
\[
K_{ij}(\mathbf{h})=
\mathcal{N}\!\big(\boldsymbol{\mu}_i(\mathbf{h});\boldsymbol{\mu}_j(\mathbf{h}),
\boldsymbol{\Sigma}_i(\mathbf{h})+\boldsymbol{\Sigma}_j(\mathbf{h})\big),
\]
and $\boldsymbol{\pi}(\mathbf{h})\in\mathbb{R}^K$ the mixture weights, we have
\[
\mathsf{H}_2\!\big(q_\phi(\cdot\mid \mathbf{h})\big)
= -\log\!\big(\boldsymbol{\pi}(\mathbf{h})^{\!\top}\mathbf{K}(\mathbf{h})\,\boldsymbol{\pi}(\mathbf{h})\big).
\]
This computation is $\mathcal{O}(K^2)$ in the number of mixture components and is negligible relative to the transformer backbone.

\section{Further Analyses}

In this section we provide additional analyses supporting observations made in the main text. Unless otherwise specified, results are reported on the TriviaQA dataset, where SSD exhibited the strongest performance variation across base models.

\subsection{Distillation Fidelity vs. Detection Performance}
\label{app:dist-fidelity}
We examine whether the quality of distillation (how closely the student learns the teacher distribution) constrains hallucination detection performance. We define \textit{distillation fidelity} ($\rho_{\text{fidelity}}$) as the Spearman rank correlation, over $N=1000$ test prompts, between the student's predicted entropy and the teacher dispersion. A high $\rho_{\text{fidelity}}$ indicates a more 
faithful recovery of the teacher's prompt-induced uncertainty ordering.

\begin{figure}[h]
  \centering
  \includegraphics[width=0.7\linewidth]{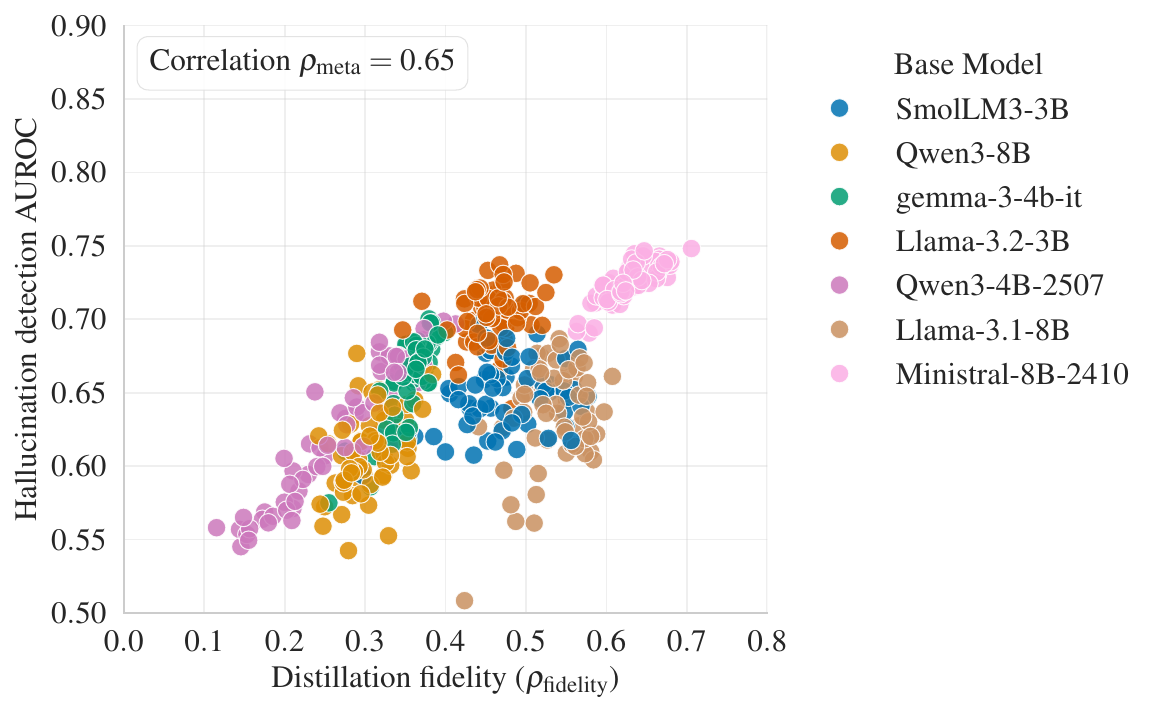}
  \caption{\textbf{Distillation fidelity drives detection performance.} The X-axis shows \textit{distillation fidelity} ($\rho_{\text{fidelity}}$): the Spearman correlation between the student's predicted entropy and the teacher dispersion across TriviaQA test prompts. Multiple points per model correspond to different student hyperparameter settings. $\rho_{\text{meta}}$ is the Spearman correlation between the X and Y variables; it shows that models where the student fails to learn the distribution (low $\rho_{\text{fidelity}}$) result in poor detection.}
  \label{fig:spearman}
\end{figure}

Figure \ref{fig:spearman} plots hallucination detection AUROC against distillation fidelity across base models and student hyperparameter settings. We observe a strong positive relationship, with a cross-model Spearman correlation of $\rho_{\text{meta}} = 0.65$, indicating that detection performance is limited by the student's ability to learn the prompt-to-dispersion mapping. The failure cases, notably Qwen3-4B, exhibit near-zero $\rho_{\text{fidelity}}$, suggesting that the relevant uncertainty signal is not accessible from the chosen prompt representation.

\subsection{Semantic Consensus from the Predicted Mixture}
\label{sec:results-sampling}

In Section~\ref{sec:distribution-uses}, we noted that modelling an explicit predictive density enables latent-space sampling and computation of mixture statistics such as the mean. In language models, the mixture mean can serve as a proxy for semantic consensus. Whereas autoregressive decoding produces a single trajectory through output space, the mixture mean aggregates across the model's sampled distribution in embedding space. We evaluate whether this analytic mean more faithfully approximates the empirical semantic centroid of sampled answers than a single low-temperature decoding.

We treat the empirical mean of $S=32$ stochastic samples in embedding space as a reference semantic centroid, then measure the mean squared distance (MSD) in Euclidean space between this centroid and both (i) the model's low-temperature output and (ii) the analytical mean of the student mixture, $\boldsymbol{\mu}_{ssd} = \sum_k \pi_k \boldsymbol{\mu}_k$. Lower MSD indicates better recovery of the model's aggregate semantic belief. MSD is computed as $\mathbb{E}_i\left[\|\hat{\mathbf{z}}_i-\boldsymbol{\mu}^{\text{true}}_i\|_2^2\right]$ over examples $i$ in the subset.

We report results for the best-performing student configuration (Ministral~8B), reflecting the strong architectural dependence of distillation fidelity. As shown in Table \ref{tab:ministral_sampling}, the SSD mixture mean provides a closer approximation to the semantic centroid than a single default decoding pass. This effect is most pronounced for incorrect answers, where SSD improves centroid recovery in over 76\% of test cases. This aligns with prior work showing that hallucinated outputs tend to deviate from semantic consensus \citep{lin2024generating, phillips2025geometricuncertaintydetectingcorrecting}. Together with the OOD results in Table \ref{tab:utility_likelihood}, this result suggests that SSD exposes a coherent semantic belief state from which both prior (entropy) and posterior (likelihood and distance-to-consensus) reliability signals can be derived.

\begin{table*}[h]
\centering
\caption{Semantic consensus approximation with Ministral 8B. We report the mean squared distance (MSD) of the default answer embedding and the SSD predicted mean to the true semantic centroid (average of $S=32$ samples). Values denote mean MSD, with the bootstrap standard deviation expressed as a percentage of the mean shown as a subscript.}
\label{tab:ministral_sampling}
\begin{tabular}{lcccc}
\toprule
Subset & Default MSD $\downarrow$ & SSD MSD $\downarrow$ & Imp. (\%) $\uparrow$ & SSD Win Rate (\%) $\uparrow$ \\
\midrule
All & 0.015\textsubscript{3.2} & \textbf{0.008\textsubscript{1.9}} & +45.5 & 58.3 \\
Correct & 0.013\textsubscript{4.4} & \textbf{0.009\textsubscript{2.3}} & +32.3 & 48.8 \\
Incorrect & 0.018\textsubscript{4.2} & \textbf{0.007\textsubscript{3.4}} & +63.0 & 76.5 \\
\bottomrule
\end{tabular}
\end{table*}

\subsection{Cross-Dataset Generalization}
\label{app:cross_dataset}

We test whether trained SSD students capture structure that transfers beyond their training distribution: a student trained on one dataset is evaluated directly on the other, with no retraining. The PCA basis fit on the source dataset is reused to project target answer embeddings, so all scores are computed in a consistent space. The teacher dispersion (TD) baseline is computed on the target dataset in the same source-fit basis, ensuring a single coordinate system for the comparison. We report results for three models, selected to span the major architectural families in our evaluation, using the evaluation protocol of Tables~\ref{tab:ed-vs-ssd} and~\ref{tab:utility_likelihood}: scores on the target test set ($N{=}1{,}000$) with bootstrap standard deviations over 1{,}000 resamples. \emph{In-dist (target)} denotes the AUROC of a student trained on the target dataset itself; the gap between this column and \emph{SSD} is the transfer cost. \emph{Mismatch} repeats the out-of-domain answer detection task of Table~\ref{tab:utility_likelihood} using the source-trained student.

\begin{table}[t]
\centering
\caption{\textbf{Cross-dataset generalization.} Students trained on the
source dataset are evaluated on the target dataset without retraining.
Values report AUROC with bootstrap standard deviation over 1{,}000
resamples of the target test set.}
\label{tab:cross_dataset}

\begin{tabular}{lcccc}
\toprule
Model & In-dist (target) & SSD & TD & Mismatch \\
\midrule
\multicolumn{5}{l}{\textbf{MMLU $\rightarrow$ TriviaQA}} \\
Qwen3 4B & 0.699 & 0.672 $\pm$ 0.017 & 0.728 $\pm$ 0.016 & 0.815 $\pm$ 0.009 \\
Llama 3.2 3B & 0.737 & 0.663 $\pm$ 0.017 & 0.664 $\pm$ 0.017 & 0.751 $\pm$ 0.011 \\
Ministral 8B & 0.748 & 0.717 $\pm$ 0.017 & 0.732 $\pm$ 0.016 & 0.780 $\pm$ 0.010 \\
\midrule
\multicolumn{5}{l}{\textbf{TriviaQA $\rightarrow$ MMLU}} \\
Qwen3 4B & 0.649 & 0.595 $\pm$ 0.018 & 0.633 $\pm$ 0.017 & 0.777 $\pm$ 0.010 \\
Llama 3.2 3B & 0.638 & 0.574 $\pm$ 0.018 & 0.669 $\pm$ 0.018 & 0.793 $\pm$ 0.010 \\
Ministral 8B & 0.635 & 0.595 $\pm$ 0.019 & 0.641 $\pm$ 0.018 & 0.838 $\pm$ 0.009 \\
\bottomrule
\end{tabular}

\end{table}

Table~\ref{tab:cross_dataset} reports the results, from which we highlight three findings. First, hallucination prediction transfers with limited cost: the mean AUROC drop relative to an in-distribution student is 0.04 (MMLU $\rightarrow$ TriviaQA) and 0.05 (TriviaQA $\rightarrow$ MMLU), comparable in both directions. Second, mismatch detection generalizes strongly (AUROC 0.75--0.84): while reduced relative to the in-distribution setting ($\approx$0.95, Table~\ref{tab:utility_likelihood}), the predicted density continues to correctly localize matched answers and reject mismatched ones under distribution shift. Third, the asymmetry between directions is mild, suggesting the result is not driven by one dataset being more `complete' than the other. Together, these results indicate that the prompt-to-density mapping captures structure that is not narrowly tied to a single training distribution.

\subsection{Richer Prompt Representations}
\label{app:richer_reps}

The main experiments condition the student on a single-token hidden state from a single model layer. Here we test whether this choice constitutes a fundamental performance ceiling by training additional students on two richer prompt
representations: \emph{concat3}, the concatenated final-token hidden states of the three deepest transformer layers, and \emph{mean10}, the element-wise mean of final-token hidden states across the deepest ten layers. For each model and representation we sweep $H \in \{128, 256, 512, 1024\}$, $D \in \{3, 6\}$, and dropout $\in \{0.2, 0.4\}$ (16 configurations), with $K{=}10$ mixture components and PCA dimension 16 fixed. We report the mean test AUROC and mean distillation fidelity $\rho_{\text{fidelity}}$ (Appendix~\ref{app:dist-fidelity}) across configurations, with standard
deviations computed over the 16 runs.

\begin{table}[t]
\centering
\caption{\textbf{Richer prompt representations.} Mean hallucination prediction AUROC and distillation fidelity over 16 student configurations per (model, representation) on TriviaQA. $\Delta$ denotes the change in mean AUROC relative to the single-layer baseline.}
\label{tab:richer_reps}

\begin{tabular}{llccc}
\toprule
Model & Prompt rep. & AUROC & $\rho_{\text{fidelity}}$ & $\Delta$ \\
\midrule
Qwen3 4B & Single layer & 0.660 $\pm$ 0.035 & 0.315 & -- \\
 & Concat3 & 0.628 $\pm$ 0.031 & 0.196 & $-0.032$ \\
 & Mean10 & 0.641 $\pm$ 0.033 & 0.251 & $-0.019$ \\
\midrule
Llama 3.2 3B & Single layer & 0.712 $\pm$ 0.015 & 0.418 & -- \\
 & Concat3 & 0.727 $\pm$ 0.016 & 0.430 & $+0.014$ \\
 & Mean10 & 0.722 $\pm$ 0.017 & 0.420 & $+0.009$ \\
\midrule
Ministral 8B & Single layer & 0.721 $\pm$ 0.027 & 0.597 & -- \\
 & Concat3 & 0.739 $\pm$ 0.019 & 0.634 & $+0.018$ \\
 & Mean10 & 0.724 $\pm$ 0.024 & 0.603 & $+0.003$ \\
\bottomrule
\end{tabular}

\end{table}

Table \ref{tab:richer_reps} shows the results. Richer representations improve both AUROC and distillation fidelity for two of the three models (Llama 3.2 3B and Ministral 8B), indicating that the single-token/single-layer representation is a conservative engineering choice rather than a fundamental limit for these models. Qwen3 4B shows the opposite pattern: both rich variants underperform the single-layer baseline, with corresponding decreases in fidelity. This is consistent with overfitting at our training set size ($N_{train}{=}4{,}000$): increasing input dimensionality without scaling supervision can hurt generalization.

Notably, Qwen3 4B also exhibits the highest sensitivity to hyperparameter choice (std 0.031--0.035 vs.\ 0.015--0.027 for the other models), itself consistent with the distillation-fidelity gap identified in Appendix~\ref{app:dist-fidelity}. We conclude that richer prompt representations are a non-trivial engineering axis whose benefit depends jointly on per-model regularization and training data quantity. A systematic exploration along the lines of \citet{kramar2026buildingproductionreadyprobesgemini} is a natural direction for future work.

\section{Ablations}
\label{sec:ablations}

We analyze two design axes underlying SSD: the dimensionality of the semantic embedding space and the capacity of the student mixture model. All ablations are conducted using the TriviaQA dataset.

\subsection{PCA Dimension}

Standard sentence embeddings are high-dimensional (e.g., $d_z=768$). Training an MDN to model a full covariance density in this space is data-intensive. We apply PCA to reduce the target space to $d_{pca} \in \{16, 32, 64, 128\}$.

\begin{figure}[h]
    \centering
    \includegraphics[width=0.6\linewidth]{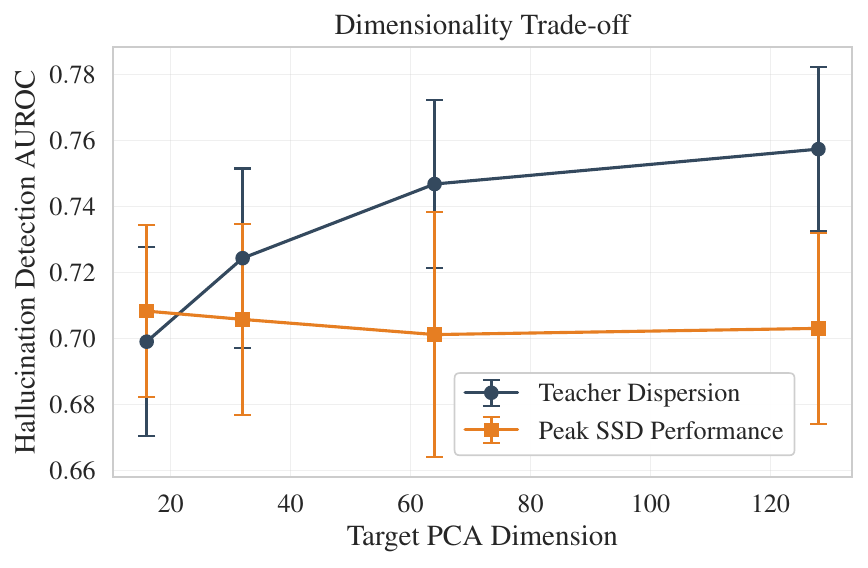}
    \caption{As PCA dimension increases, the semantic resolution improves and the teacher dispersion baseline rises. The distillation task however becomes harder, causing SSD performance to fall.}
    \label{fig:pca_tradeoff}
\end{figure}

This introduces a trade-off between semantic resolution and distillation fidelity: higher dimensions preserve more semantic detail (raising the teacher dispersion baseline), but make the prompt-to-density regression task harder for the student to learn with fixed data.
As shown in Figure \ref{fig:pca_tradeoff} and Table \ref{tab:pca_ablation}, while the detection performance of teacher dispersion improves monotonically with dimensionality, the student's performance (SSD) peaks at $d_{pca}=16$. Beyond this point, the student fails to effectively distil the increasingly sparse density, leading to a drop in detection accuracy. For the OOD experiments we find PCA dimensions of 16 and 32 perform similarly, followed by a drop-off at higher dimensions. We fix $d_{pca}=16$ for all main experiments bar the OOD task, where we set $d_{pca}=32$.

We underline that the same training dataset size ($N_{train}=4000$) was used in all PCA ablations; the learning task is therefore more difficult in higher dimensions, as models must learn more complex relationships using the same number of datapoints. For larger training datasets, we expect the SSD performance will continue to match or exceed the TD performance.

\begin{table}[h]
\centering
\caption{Effect of PCA dimensionality on Hallucination prediction (SSD) and OOD Verification. Scores are averaged across the best student configuration for each model.}
\label{tab:pca_ablation}
\begin{tabular}{cccc}
\toprule
PCA Dim & Teacher Dispersion (TD) & Peak SSD AUROC & Peak OOD AUROC \\
\midrule
16 & 0.699 $\pm$ 0.029 & \textbf{0.708 $\pm$ 0.026} & 0.965 $\pm$ 0.026 \\
32 & 0.724 $\pm$ 0.027 & 0.706 $\pm$ 0.029 & \textbf{0.966 $\pm$ 0.035} \\
64 & 0.747 $\pm$ 0.025 & 0.701 $\pm$ 0.037 & 0.960 $\pm$ 0.047 \\
128 & \textbf{0.757 $\pm$ 0.025} & 0.703 $\pm$ 0.029 & 0.950 $\pm$ 0.055 \\
\bottomrule
\end{tabular}
\end{table}

\subsection{Student Capacity} 
We investigate whether the performance gap on difficult models can be closed by increasing the student's capacity. We analyze two axes of complexity:
\begin{enumerate}
    \item \textbf{Distributional Complexity:} We vary the number of mixture components $K \in \{1, 2, 5, 10\}$.
    \item \textbf{Backbone Capacity:} We vary the MLP hidden dimension $H \in \{128, 256, 512, 1024\}$ and depth $D \in \{2, 3, 4, 6\}$.
\end{enumerate}

\begin{figure}[h]
    \centering
    \includegraphics[width=\linewidth]{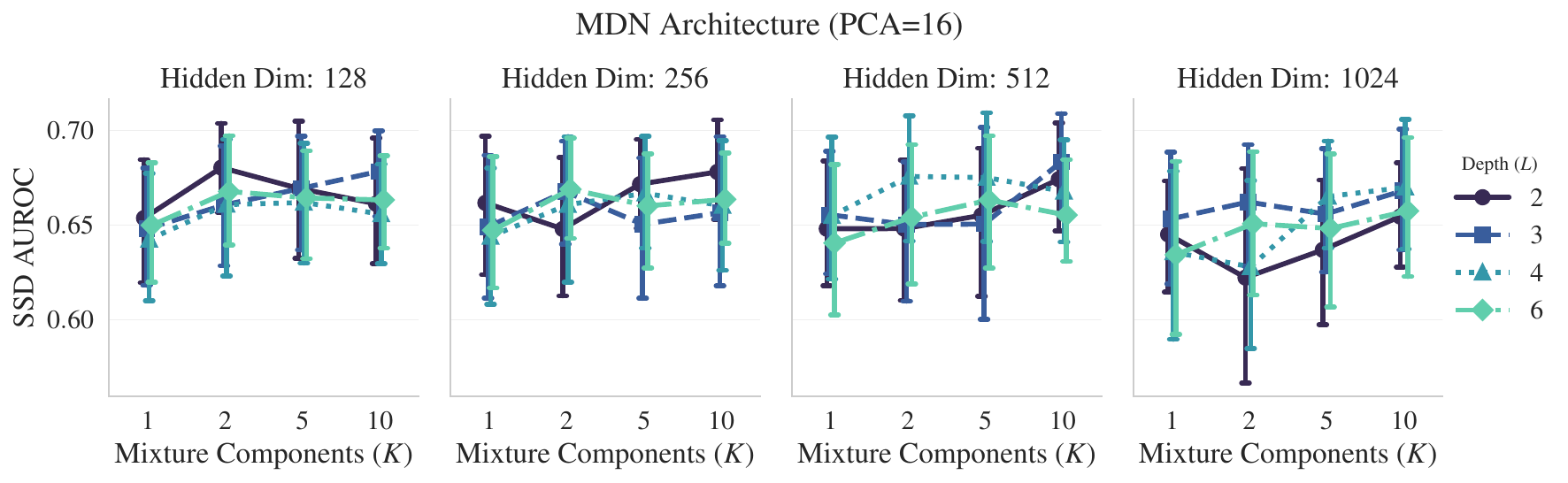}
    \caption{Student capacity ablation on the TriviaQA dataset, averaged across all seven models investigated.}
    \label{fig:capacity-ablation}
\end{figure}

Table \ref{tab:metadata} reveals that the optimal hyperparameters vary by base model. Aggregating results across all models (Table \ref{tab:marginal_ablation}) reveals \textbf{capacity saturation}, where increasing the network width or depth yields negligible gains in mean AUROC. Figure~\ref{fig:capacity-ablation} visualizes the full sweep across width,
depth, and mixture components. This saturation indicates that performance is not limited by student expressivity, but by the information available in the prompt representation used for distillation. The student appears to extract most of the accessible uncertainty signal from this representation; further improvements would require richer inputs (e.g., attention-pooled representations) rather than larger students. Although we observe optimal performance for $K=10$ mixture components, we opt not to increase this further given our set number of $S=32$ samples; modelling more than ten unique semantic concepts in this setting is unrealistic and would likely damage the generalization ability of the trained student.

\begin{table}[h]
\caption{Marginal impact of student capacity (Fixed PCA=16). Scores represent the mean AUROC across all runs with the specified parameter; standard deviations indicate variability across different base models and remaining hyperparameter configurations. We observe negligible performance gains from increasing complexity, indicating information saturation in the prompt representation.}
\label{tab:marginal_ablation}
\centering
\begin{subtable}{0.32\linewidth}
\centering
\caption{Components ($K$)}
\begin{tabular}{cc}
\toprule
$K$ & AUROC \\
\midrule
1 & 0.648 $\pm$ 0.050 \\
2 & 0.657 $\pm$ 0.052 \\
5 & 0.660 $\pm$ 0.048 \\
10 & \textbf{0.666 $\pm$ 0.042} \\
\bottomrule
\end{tabular}
\end{subtable}
\hfill
\begin{subtable}{0.32\linewidth}
\centering
\caption{Width ($H$)}
\begin{tabular}{cc}
\toprule
$H$ & AUROC \\
\midrule
128 & \textbf{0.662 $\pm$ 0.043} \\
256 & 0.660 $\pm$ 0.046 \\
512 & 0.660 $\pm$ 0.050 \\
1024 & 0.649 $\pm$ 0.054 \\
\bottomrule
\end{tabular}
\end{subtable}
\hfill
\begin{subtable}{0.32\linewidth}
\centering
\caption{Depth ($D$)}
\begin{tabular}{cc}
\toprule
$L$ & AUROC \\
\midrule
2 & 0.657 $\pm$ 0.049 \\
3 & \textbf{0.660 $\pm$ 0.047} \\
4 & 0.658 $\pm$ 0.050 \\
6 & 0.656 $\pm$ 0.048 \\
\bottomrule
\end{tabular}
\end{subtable}
\end{table}

\section{Model Configuration and Selection}
\label{sec:model-selection}

Here we summarize model characteristics and specific configuration choices used throughout our experiments. Table~\ref{tab:metadata} reports base model accuracy on TriviaQA, the optimal probe layers identified for PCP and SEP, and the student architecture that achieved the highest SSD AUROC for each model.

\begin{table}[h]
\centering
\caption{Model-specific configuration details on TriviaQA. `Accuracy' denotes base model QA accuracy. `Total' gives the number of transformer layers, while `PCP' and `SEP' report the selected probe layers. `Best SSD Config' lists the student architecture ($H$=hidden size, $D$=depth, $K$=mixture components) achieving the highest SSD AUROC.}
\label{tab:metadata}
\begin{tabular}{l c ccc | cccc}
\toprule
 & & \multicolumn{3}{c}{Layers} & \multicolumn{4}{c}{Best SSD Config} \\
Model & Accuracy & Total & PCP & SEP & H & D & K & AUROC \\
\midrule
Qwen3-8B & 0.620 & 36 & 32 & 24 & 128 & 6 & 10 & 0.677 \\
Qwen3-4B & 0.565 & 36 & 18 & 20 & 256 & 2 & 10 & 0.699 \\
Llama 3.1 8B & 0.700 & 32 & 20 & 16 & 256 & 4 & 2 & 0.687 \\
Llama 3.2 3B & 0.585 & 28 & 15 & 19 & 512 & 3 & 5 & 0.737 \\
Ministral 8B & 0.656 & 36 & 33 & 34 & 1024 & 6 & 1 & 0.748 \\
SmolLM3 3B & 0.567 & 36 & 20 & 19 & 512 & 6 & 5 & 0.710 \\
Gemma 3 4B & 0.532 & 34 & 22 & 19 & 512 & 2 & 10 & 0.700 \\
\bottomrule
\end{tabular}
\end{table}

\subsection{Layer Selection Protocol}
\label{sec:layer-selection}

To select the student input layer $\ell^*$, we follow \citet{kossen2024semanticentropyprobesrobust} and train linear probes on all layers to predict binarized Semantic Entropy. We choose the layer that maximizes validation accuracy on this proxy task. This heuristic assumes that layers most predictive of scalar dispersion also contain the richest signal for learning the full semantic density. In practice, the selected layers tend to lie in deeper parts of the network (Table~\ref{tab:metadata}).

\section{Computational Complexity Analysis}
\label{app:complexity}

The primary motivation for Semantic Self-Distillation is to enable generic uncertainty estimation in AR models without sampling. Here we compare the asymptotic complexity of our method against sampling-based baselines. Let $T_{\text{pre}}$ be the prompt length, $T_{\text{gen}}$ be the generation length, $S$ be the number of samples, and $C_{\text{LLM}}$ be the cost of a single transformer forward pass per token.

\paragraph{Sampling-based Methods (SE, TD).} 
Computing Semantic Entropy \citep{farquhar_detecting_2024} requires $S$ stochastic sequences per prompt at inference time (typically $S \in [5, 20]$). The cost is dominated by the autoregressive decoding steps:
\[ 
\mathcal{O}_{\text{SE}} \approx S \times (T_{\text{pre}} + T_{\text{gen}}) \times C_{\text{LLM}} + \mathcal{O}_{\text{NLI}}(S^2)
\]
where $\mathcal{O}_{\text{NLI}}$ represents the quadratic cost of comparing all sample pairs using an entailment model. This linear scaling with $S$ introduces latency that is unacceptable in many applications, for example in interactive agentic systems.

\paragraph{Scalar Probes (PCP, SEP).}
Scalar probes operate on prompt representations extracted from a single forward pass of the base model. The probe itself is a lightweight classifier or regressor applied to a cached hidden state, without requiring additional sampling or decoding:
\[
\mathcal{O}_{\text{SEP}} \approx 1 \times T_{\text{pre}} \times C_{\text{LLM}} + C_{\text{MLP}}
\]
While efficient, these methods predict a compressed scalar statistic and discard the geometric structure of the model's semantic output distribution.

\paragraph{Semantic Self-Distillation (SSD).}
SSD matches the inference complexity of scalar probes while retaining the distributional richness of sampling methods. The MDN head introduces a slightly larger parameter set than a linear probe, but this cost ($C_{\text{MDN}}$) is negligible compared to the transformer backbone. Crucially, the entropy calculation is analytical and does not require sampling:
\[ 
\mathcal{O}_{\text{SSD}} \approx 1 \times T_{\text{pre}} \times C_{\text{LLM}} + C_{\text{MDN}}
\]
Thus, SSD effectively moves the computational burden from \textit{inference time} (where latency matters) to \textit{training time}, enabling $O(1)$ uncertainty estimation during deployment.


\section{Full Hallucination Prediction Results}
\label{app:full-hd-results}

We report full per-model hallucination detection results corresponding to the summaries in Tables~\ref{tab:ed-vs-ssd} and~\ref{tab:method-attribs}. These tables provide AUROC and AUPRC metrics for all methods across datasets.

\begin{table}[h]
\centering 
\caption{\textbf{Hallucination Prediction AUROC.}
Values report mean AUROC (\%) over 1{,}000 bootstrap resamples of the test set, with the bootstrap standard deviation shown as a subscript.
We mark the best method per dataset and model in bold.}
\label{tab:main_results}
\begin{tabular}{lccccc|ccccc}
\toprule
 & \multicolumn{5}{c}{TriviaQA} & \multicolumn{5}{c}{MMLU} \\
\cmidrule(lr){2-6} \cmidrule(lr){7-11}
Model & PCP & SEP & SE & TD & SSD & PCP & SEP & SE & TD & SSD \\
\midrule
Qwen3 8B & 76.9\textsubscript{1.5} & 76.2\textsubscript{1.5} & \textbf{81.6\textsubscript{1.4}} & 68.3\textsubscript{1.6} & 67.7\textsubscript{1.8} & \textbf{69.8\textsubscript{1.6}} & 59.5\textsubscript{1.8} & 63.7\textsubscript{1.6} & 63.5\textsubscript{1.7} & 57.1\textsubscript{1.8} \\
Qwen3 4B & \textbf{80.2\textsubscript{1.4}} & 78.3\textsubscript{1.4} & 78.6\textsubscript{1.4} & 73.2\textsubscript{1.6} & 69.9\textsubscript{1.6} & \textbf{67.9\textsubscript{1.7}} & 64.3\textsubscript{1.8} & 60.5\textsubscript{1.6} & 63.1\textsubscript{1.8} & 64.9\textsubscript{1.7} \\
Llama 3.1 8B & 77.0\textsubscript{1.7} & 74.4\textsubscript{1.7} & \textbf{85.1\textsubscript{1.3}} & 69.6\textsubscript{1.7} & 68.7\textsubscript{1.9} & \textbf{71.5\textsubscript{1.6}} & 64.9\textsubscript{1.7} & 70.6\textsubscript{1.6} & 69.0\textsubscript{1.6} & 63.8\textsubscript{1.7} \\
Llama 3.2 3B & 76.7\textsubscript{1.5} & 75.7\textsubscript{1.6} & \textbf{80.6\textsubscript{1.4}} & 66.6\textsubscript{1.7} & 73.7\textsubscript{1.6} & 66.0\textsubscript{1.9} & 64.4\textsubscript{1.8} & \textbf{70.1\textsubscript{1.7}} & 67.1\textsubscript{1.8} & 63.8\textsubscript{1.7} \\
Ministral 8B & 74.0\textsubscript{1.7} & 77.1\textsubscript{1.6} & \textbf{80.5\textsubscript{1.5}} & 74.3\textsubscript{1.7} & 74.9\textsubscript{1.6} & 67.0\textsubscript{1.7} & 62.8\textsubscript{1.8} & \textbf{69.3\textsubscript{1.7}} & 65.1\textsubscript{1.8} & 63.6\textsubscript{1.8} \\
SmolLM3 3B & 79.6\textsubscript{1.4} & 78.7\textsubscript{1.4} & \textbf{83.1\textsubscript{1.2}} & 69.9\textsubscript{1.6} & 71.1\textsubscript{1.6} & 67.0\textsubscript{1.7} & 66.3\textsubscript{1.8} & \textbf{68.8\textsubscript{1.7}} & 62.3\textsubscript{1.8} & 62.0\textsubscript{1.8} \\
Gemma 3 4B & \textbf{75.2\textsubscript{1.6}} & 70.6\textsubscript{1.6} & 73.1\textsubscript{1.5} & 67.6\textsubscript{1.6} & 70.0\textsubscript{1.7} & \textbf{66.2\textsubscript{1.8}} & 57.8\textsubscript{1.8} & 61.3\textsubscript{1.6} & 62.8\textsubscript{1.7} & 60.7\textsubscript{1.8} \\
\bottomrule
\end{tabular}
\end{table}

\begin{table}[h]
\centering
\caption{\textbf{Hallucination Prediction AUPRC.}
Values report mean AUPRC (\%) over 1{,}000 bootstrap resamples of the test set, with the bootstrap standard deviation shown as a subscript. We mark the best method per dataset and model in bold.}
\label{tab:auprc_results}
\begin{tabular}{lccccc|ccccc}
\toprule
 & \multicolumn{5}{c}{TriviaQA} & \multicolumn{5}{c}{MMLU} \\
\cmidrule(lr){2-6} \cmidrule(lr){7-11}
Model & PCP & SEP & SE & TD & SSD & PCP & SEP & SE & TD & SSD \\
\midrule
Qwen3 8B & 64.8\textsubscript{2.6} & 64.9\textsubscript{2.7} & \textbf{75.0\textsubscript{2.1}} & 52.4\textsubscript{2.6} & 56.9\textsubscript{2.8} & \textbf{66.9\textsubscript{2.3}} & 56.3\textsubscript{2.3} & 60.3\textsubscript{2.2} & 61.9\textsubscript{2.3} & 55.9\textsubscript{2.3} \\
Qwen3 4B & \textbf{74.3\textsubscript{2.4}} & 70.8\textsubscript{2.5} & 73.4\textsubscript{2.2} & 59.7\textsubscript{2.4} & 61.9\textsubscript{2.5} & \textbf{60.1\textsubscript{2.5}} & 57.6\textsubscript{2.5} & 53.4\textsubscript{2.3} & 58.5\textsubscript{2.4} & 59.4\textsubscript{2.5} \\
Llama 3.1 8B & 59.7\textsubscript{3.1} & 54.0\textsubscript{3.0} & \textbf{73.4\textsubscript{2.7}} & 44.4\textsubscript{2.7} & 51.9\textsubscript{2.9} & \textbf{72.7\textsubscript{2.1}} & 62.6\textsubscript{2.3} & 70.5\textsubscript{2.1} & 70.9\textsubscript{2.1} & 62.7\textsubscript{2.4} \\
Llama 3.2 3B & 68.9\textsubscript{2.3} & 65.4\textsubscript{2.6} & \textbf{74.1\textsubscript{2.2}} & 55.8\textsubscript{2.6} & 65.5\textsubscript{2.5} & 76.4\textsubscript{1.9} & 74.5\textsubscript{2.0} & \textbf{78.4\textsubscript{1.8}} & 74.0\textsubscript{2.0} & 74.5\textsubscript{1.9} \\
Ministral 8B & 59.2\textsubscript{3.0} & 62.9\textsubscript{2.8} & \textbf{69.5\textsubscript{2.8}} & 57.9\textsubscript{2.8} & 59.9\textsubscript{2.8} & 72.9\textsubscript{2.2} & 70.7\textsubscript{2.1} & \textbf{76.0\textsubscript{1.9}} & 71.4\textsubscript{2.1} & 71.7\textsubscript{2.1} \\
SmolLM3 3B & 75.0\textsubscript{2.1} & 73.1\textsubscript{2.2} & \textbf{78.8\textsubscript{1.9}} & 57.1\textsubscript{2.5} & 62.4\textsubscript{2.5} & 67.7\textsubscript{2.3} & 67.6\textsubscript{2.2} & \textbf{70.5\textsubscript{2.2}} & 64.6\textsubscript{2.2} & 63.0\textsubscript{2.3} \\
Gemma 3 4B & \textbf{69.2\textsubscript{2.5}} & 64.7\textsubscript{2.5} & 69.2\textsubscript{2.2} & 63.1\textsubscript{2.3} & 66.1\textsubscript{2.5} & \textbf{73.5\textsubscript{2.1}} & 63.8\textsubscript{2.1} & 67.6\textsubscript{1.9} & 71.1\textsubscript{2.0} & 69.2\textsubscript{2.1} \\
\bottomrule
\end{tabular}
\end{table}

\end{document}